%% file: main.tex
\documentclass[letterpaper, 10 pt, conference]{ieeeconf}
\IEEEoverridecommandlockouts
\usepackage{cite}
\usepackage{calc}
\usepackage{float}
\usepackage{amsmath,amssymb,amsfonts}
\usepackage{algorithmic}
\usepackage{graphicx}
\usepackage{textcomp}
\usepackage{xcolor}
\usepackage{booktabs}
\usepackage{hyperref}  
\usepackage{cleveref}
\usepackage{pgfplots}
\pgfplotsset{compat=1.18}
\usepackage{tikz}
\usepackage{subcaption}
\usepackage{adjustbox} 
\usepackage{etoolbox}   
\usetikzlibrary{arrows.meta,positioning,fit,shapes.misc}
\def\BibTeX{{\rm B\kern-.05em{\sc i\kern-.025em b}\kern-.08em
    T\kern-.1667em\lower.7ex\hbox{E}\kern-.125emX}}

\begin{document}

\title{\LARGE \bf{World Model Failure Classification and Anomaly Detection for Autonomous Inspection\\
}}
\author{Michelle Ho,$^{1,2}$ 
Muhammad Fadhil Ginting$^{1,2}$, 
Isaac R. Ward$^{1}$, 
Andrzej Reinke$^2$,
Mykel J. Kochenderfer$^1$,  \\
Ali-akbar Agha-Mohammadi$^2$, 
and Shayegan Omidshafiei$^2$
\thanks{$^{1}$Stanford University, Stanford, CA 94305 USA
{\tt\small \{mtho, ginting, irward, mykel\}@stanford.edu}}%
\thanks{$^{2}$Field AI, Mission Viejo, CA 92691 USA}%
}

\maketitle

\begin{abstract}
Autonomous inspection robots for monitoring industrial sites can reduce costs and risks associated with human-led inspection. However, accurate readings can be challenging due to occlusions, limited viewpoints, or unexpected environmental conditions. We propose a hybrid framework that combines supervised failure classification with anomaly detection, enabling classification of inspection tasks as a success, known failure, or anomaly (i.e., out-of-distribution) case. Our approach uses a world model backbone with compressed video inputs. This policy-agnostic, distribution-free framework determines classifications based on two decision functions set by conformal prediction (CP) thresholds before a human observer does. We evaluate the framework on gauge inspection feeds collected from office and industrial sites and demonstrate real-time deployment on a Boston Dynamics Spot. Experiments show over 90\% accuracy in distinguishing between successes, failures, and OOD cases, with classifications occurring earlier than a human observer. These results highlight the potential for robust, anticipatory failure detection in autonomous inspection tasks or as a feedback signal for model training to assess and improve the quality of training data. Project website: \url{https://autoinspection-classification.github.io/}
\end{abstract}

\section{Introduction}
\input{introduction}

\section{Related Work}
\input{related-work}

\section{Methodology} 
\input{formulation}
\input{methodology}

\section{Experimental Results}
\input{results}

\section{Conclusion}
\input{conclusion}

\bibliographystyle{IEEEtran}
\bibliography{references.bib}

\subsection{Acknowledgements}

Generative AI tools and technologies were used in this work solely for validating ideas in the brainstorming stage (ChatGPT),  tab-autocompleting while extending prewritten code to new metrics, with all methods and analyses otherwise unchanged (Cursor using Claude), and for figure placement and table creation in LaTeX (ChatGPT). 

\appendix
\subsection{Low Resolution Results}
We ran additional experiments to observe if higher image compression before the world model would affect results. \Cref{tab:lowres-accuracy} depicts the results from 50\% higher compression.

\input{tables/lowres_accuracy}

\subsection{Reproducibility}
The code and subset of the data required to reproduce these results are available at \url{https://autoinspection-classification.github.io/}. Please see below for the GPU architecture and training configurations for the models.

\begin{table}[h]
\centering
\caption{Model Training Configurations}
\label{tab:train_params}
\begin{tabular}{ll}
\hline
Optimizer & AdamW with weight decay ($10^{-4}$) \\
Batch size & Default 32 (configurable) \\
Learning rate & Defined in hyperparameters \\
Gradient clipping & Value of 1.0 \\
Early stopping & Patience 15 epochs, $\Delta = 10^{-5}$ \\
Check-pointing & Minimum validation loss \\
Epochs & Configurable (default: 10) \\
Device selection & GPU with CUDA, fallback to CPU \\
Data workers & Training: 3, Validation: 1 (default) \\
\hline
\end{tabular}
\end{table}

All simulations were executed on an g4dn.xlarge Ubuntu EC2 Amazon Web Services Instance with 100 GiB of storage, 16 GiB RAM, and 4 CPU cores.

\end{document}

%% file: introduction.tex
Industrial inspection robots can offer consistent, reliable monitoring of instruments in large facilities, reducing costs and inconsistencies from human-led inspections \cite{bostondynamics2024three}. In these environments, accurate readings are critical. However, vision may be limited when instruments are partially occluded by other equipment or truncated in the field of view. Additional challenges arise due to environmental conditions such as glare and shadows, which likely fall outside of the known cases for which the robot was trained. To successfully complete its mission, the robot needs to not only get accurate instrument readings but also detect and react appropriately to these perception failures. In this work, we address the challenge of accurate, efficient online failure detection for an inspection robot, focusing on both known failure cases, such as systemic misreads, and anomalous, out-of-distribution (OOD) cases due to environmental conditions. 

Human-led inspections are costly, unreliable, and complex to scale up. Mobile robots address these limitations by providing reliable measurements on a regular schedule. They enable inexpensive, time-efficient inspection without human fatigue and reduce unnecessary exposure in hazardous domains \cite{shukla2016application}. Legged robots have been tested in offshore high-voltage environments \cite{gehring2021anymal} and in uncertain domains requiring risk-aware planning and semantic mapping \cite{ginting2024seek}. These mobile robots have been adopted by industry players for automating inspection tasks such as thermal and acoustic monitoring, 3D mapping of facilities, and gauge inspection~\cite{fieldai2025technology, energy2025automate, anybotics2025automate}.

Greater reliance on autonomy introduces new risks. For instance, if a robot misses a dangerous reading on a high-pressure system, an anomaly can escalate into a major safety risk. Fault and failure detection have long histories in process systems \cite{gao2015survey}. Still, modern inspection robots introduce additional challenges due to learned perception and operation in unseen environments. Effective autonomous inspection requires detection methods that can identify known failures but also adapt to novel conditions not encountered during training. To address these challenges, OOD detection provides early warnings when inputs deviate from training distributions \cite{chandola2009anomaly}. Recent work demonstrates that it is possible to detect failures without explicit labeled failure data with conformal prediction thresholding \cite{xu_can_2025, ward2025world}. However, OOD detection alone is insufficient, as it does not provide informative actions to take upon detection, unlike approaches that leverage knowledge of recognized failure cases.

\begin{figure}[t]
    \centering
    \includegraphics[width=0.99\linewidth]{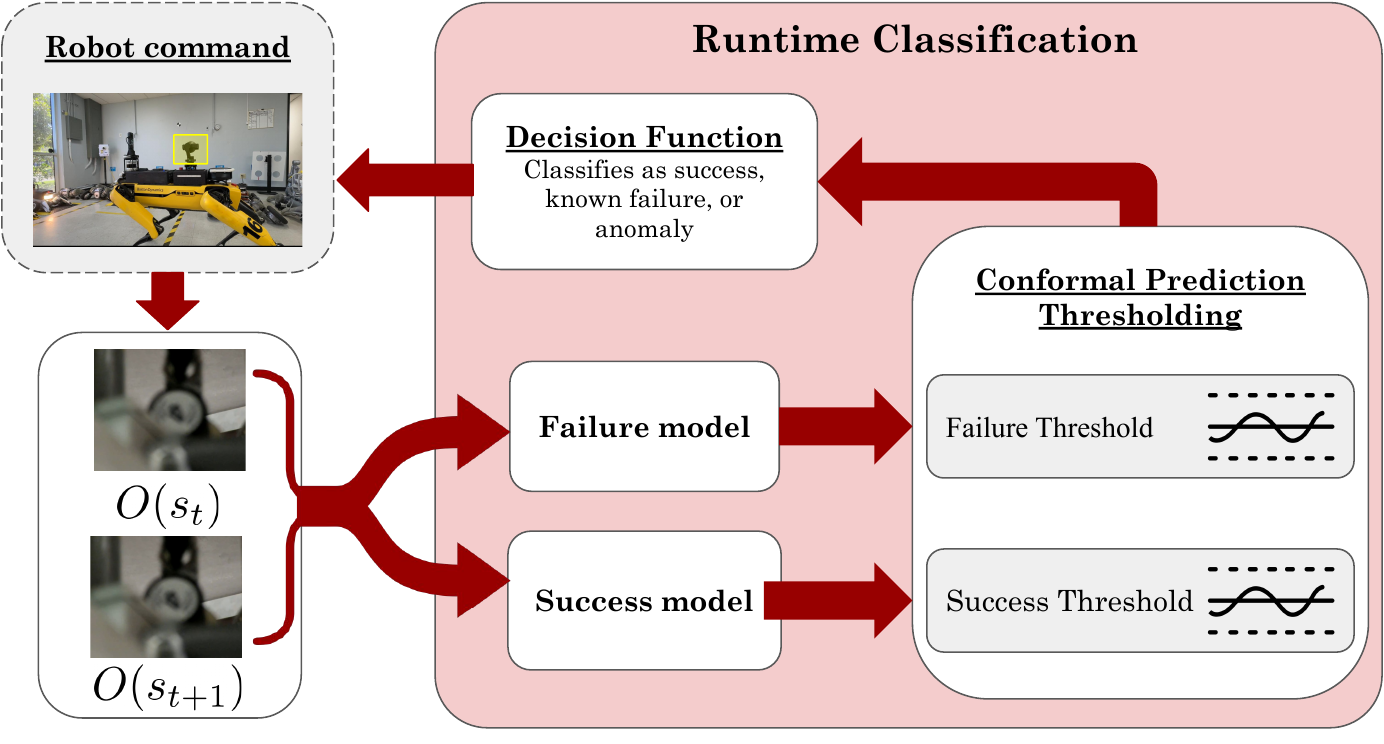}
    \caption{Our framework classifies successes, known failures, and anomalies with a world model backbone and conformal prediction thresholding, allowing for the robot to take informative actions based on the classification.}
    \label{fig:runtime_classification}
\end{figure}

In this paper, we present a failure detector for real-time anomaly monitoring for an autonomous gauge inspection problem (see \Cref{fig:runtime_classification}). Our method combines supervised failure classification with sequential OOD detection through conformal prediction–based thresholding. The detector builds on previous work that uses a world-model backbone to predict observations, from which anomaly scores are derived with prediction errors~\cite{ward2025world}. Unlike prior work that focused solely on using success data only to classify anomalies \cite{xu_can_2025, ward2025world}, we incorporate explicit classification of known failure cases. We make the following technical contributions:\\
\begin{enumerate}
    \item A hybrid detector that combines failure classification, for providing informed, corrective actions, with OOD detection, for failures not seen during training;
    \item Calibrated conformal prediction thresholds based on standard CP metrics, enabling classification of successes, known failures, and OOD readings;
    \item Evaluation on gauge inspection data gathered in both office and industrial settings;
    \item Real-time, online deployment of the detector in a gauge inspection scenario on a Boston Dynamics Spot robot.
\end{enumerate}

These contributions not only enable robust real-time failure detection but can also assess training data quality to guide targeted data collection for large-scale model training.
















%% file: related-work.tex
\subsection{Out-of-Distribution Detection}
Detecting anomalies is critical for safe, robust autonomous systems, especially in safety-critical domains such as industrial inspection. Anomalies are often defined as out-of-distribution from what the robot has seen during training. OOD detection flags anomalies, since they could indicate possible failure modes~\cite{chandola2009anomaly}. Approaches to OOD detection are vast. Statistical and reconstruction methods can be simple and interpretable but struggle with high dimensionality, temporal dependencies, or noisy data~\cite{chandola2009anomaly,hastie2009elements,nitsch2021out}. Embedding-based~\cite{lee2018simple,defard2021padim}, distribution modeling \cite{xiao2020likelihood,wang2020further,izmailov2020semi-supervised,ward2025improving,nalisnick2019deep}, ensemble~\cite{lakshminarayanan2017simple}, and Bayesian approaches~\cite{blundell2015weight} scale better to complex settings but require large, diverse datasets.

Recent work has focused on detecting failures without failure data. One method, FAIL-Detect\cite{xu_can_2025}, learns scalar signals correlated with failures and calibrates them with time-varying conformal prediction thresholds~\cite{xu_can_2025}. Another recent work builds on this method by using a world model backbone to detect failures in a highly dimensional problem. It also incorporates history to anticipate future failures before they occur \cite{ward2025world}. Though these methods do not require labeled data for training, in many cases, failure data can be readily collected during the training process and could be used to improve performance. 

\subsection{Failure Classification}
While anomaly detection identifies deviations from nominal behavior, failure classification can categorize these deviations if they have been seen before. For industrial inspection, these failure classes can include mobility and actuation~\cite{aalobaidy2020faults,gao2015survey}, perception~\cite{daftry2016introspective}, and task-level planning and execution~\cite{lussier2007fault}. Residual-based approaches for classification are interpretable and leverage a system model to generate diagnostic signals, but they are sensitive to noise or approximately modeled dynamics~\cite{basha2018multivariate, hofmann2020hidden}. Supervised learning methods are useful when the model is not easily defined~\cite{maincer2022fault, costa2019failure}. Large language models provide semantic understanding of the scene beyond recognizing pixel patterns~\cite{elhafsi2023semantic,sinha2024real-time}. Hybrid approaches for detecting different known failure cases include adding a supervised head to an autoencoder~\cite{oza2019c2ae} or combining separate frameworks for detecting various types of failures~\cite{agia2024unpacking}. In contrast, our approach detects multiple categories of perception failures using a single pretrained world model.

Though failure detection is useful for prescribing informed, corrective actions for known cases, it is not possible to train a robot to recognize every future failure it may ever encounter. Therefore, we address this by creating a unified framework that combines failure classification with anomaly detection. 

\subsection{Hybrid Anomaly Detection and Failure Classification}
Some anomaly detection methods can be combined with failure classification when labeled data for specific cases is available. For instance, one approach combines principal component analysis with supervised classifiers \cite{ding2008residual}, another adds classification heads to autoencoders \cite{oza2019c2ae}, and another uses ensembles for detecting anomaly signals and making class predictions~\cite{lakshminarayanan2017simple}. In contrast, we combine anomaly detection and failure classification using a single world model backbone to reduce architecture complexity for deployment.  

Moreover, this combination not only supports timely, more informed detection but also supports intervention strategies for specific failure modes. Some prior work focuses on defining failure beyond collision and overriding actions to prevent failure, such as with latent space binary failure classification using Hamilton–Jacobi reachability~\cite{nakamura2025generalizing}. However, our focus on combining anomaly detection with multi-class failure classification not only flags unseen deviations but also distinguishes among known failure modes, enabling more informative responses. We test our framework on a real-time hardware setup and trigger appropriate intervention strategies given the classification. 

\subsection{Conformal Prediction}
Conformal prediction (CP) is a statistical framework that quantifies uncertainty by calibrating thresholds \cite{angelopoulos2023conformal}. For a trained model and a calibration dataset, CP produces prediction intervals that contain the true label (or nominal behavior) by some user-specified metric \cite{angelopoulos2023conformal}. Since CP can be applied to black-box models, it is beneficial for safety-critical robotic applications. It is common to separate available data into distinct training and calibration sets~\cite{zhou2025conformal}. Typical scoring methods include probability-based scores (e.g., margin score, inverse probability score, negative log-likelihood), residual-based scores (e.g., regression residuals, reconstruction error), and distance scores (e.g., Mahalanobis distance, latent standard deviation, clustering distance) \cite{zhou2025conformal}. Other scoring methods require policy outputs or perform better with an exocentric view \cite{xu_can_2025}. We do not consider these in our setting because our inspection pipeline operates on robot-egocentric visual feeds for a black-box policy.

CP is typically used to construct prediction sets with some guaranteed coverage. For anomaly detection, CP can be applied to calibrate thresholds on scalar scores, setting approximate bounds that separate nominal and OOD inputs rather than interval coverage. In line with recent work on failure detection~\cite{xu_can_2025, ward2025world}, we adopt this thresholding method, applying CP bands to data gathered from a camera feed. We are not aware of prior work in robotics that uses CP in a hybrid manner to both classify failure modes and reject out-of-distribution inputs within a single framework.

%% file: formulation.tex

\subsection{Problem Formulation}
Our goal is to determine if and when a robot will fail its task and to distinguish between a known failure and an anomaly. For a given task, we represent the robot’s trajectory over a finite horizon $T$ as the sequence $\tau_T = \{s_0, a_0, \ldots, s_T, a_T\}$. The objective is to determine whether there exists a failure time $k \leq T$, and if so, to identify $k$. 

We assume an observation function $O(s_t)$ that is noise-free and fully observable, such that observations provide a direct mapping to the underlying state and thereby reveal whether the robot is failing its task. The policy $\pi(s_t) = a_t$ is tele-operated, making it unknown to the framework. Therefore, we work with the observable trajectory $\tau_t \triangleq \{o_0,\ldots,o_t\}$ which is fully informative about the underlying states. 

Like in Xu et al. \cite{xu_can_2025}, we also frame our detector as a decision function $D(\tau_t;\theta)$, where $\theta$ contains the function parameters. This function decides whether a failure (1) occurs or not (0) at time $t$ in the trajectory. An ideal detector minimizes the delay between the true failure time $k$ and the detected time $\hat{k}$.

\subsection{Failure Classification and Anomaly Detection Framework}
Given observation trajectories, our full decision function is comprised of two decision functions, $D_{\text{success}}(\tau_t; \theta_s)$ and $D_{\text{fail}}(\tau_t; \theta_f)$ parameterized by $\theta_s$ and $\theta_f$ respectively. A trajectory is classified as OOD only if both decision functions return $1$, i.e.,
\begin{equation}
    D_{\text{OOD}}(\tau_t) = \mathbf{1}\big( D_{\text{success}}(\tau_t; \theta_s) = 1 \ \wedge \ D_{\text{fail}}(\tau_t; \theta_f) = 1 \big).
    \label{eq:total_decision_fxn}
\end{equation}
Under this formulation, our detected time is 
\begin{equation}
    \hat{k} = \min \left\{ t \leq T \;:\; D_{\text{success}}(\tau_t; \theta_s) = 1 \;\lor\; D_{\text{fail}}(\tau_t; \theta_f) = 1 \right\}.
\end{equation}

We label the classification at $\hat{k}$ by
\begin{equation}
    \hat{c}(\tau_{\hat{k}})=
    \begin{cases}
        \text{success}, & 
        \begin{aligned}[t]
            &\text{if } D_{\text{success}}(\tau_{\hat{k}};\theta_s)=0 \\
            &\wedge\; D_{\text{fail}}(\tau_{\hat{k}};\theta_f)=1,
        \end{aligned}\\[4pt]
        \text{known failure}, & 
        \begin{aligned}[t]
            &\text{if } D_{\text{fail}}(\tau_{\hat{k}};\theta_f)=0 \\
            &\wedge\; D_{\text{success}}(\tau_{\hat{k}};\theta_s)=1, \\
        \end{aligned} \\
         \text{anomalous}, & \text{if } D_{\text{OOD}}(\tau_{\hat{k}})=1.\\[4pt]
    \end{cases}
\end{equation}
\Cref{fig:decision_function} shows the decision function.

\input{figures/decision_function}

The two underlying decision functions are designed with the following framework: 
\begin{enumerate}
    \item Train two predictive models on observation pairs: 
    $M_{\text{success}}(\cdot;\phi_s)$ using successful data and 
    $M_{\text{fail}}(\cdot;\phi_f)$ using failure data. 

    \ \item For a set of success and failure calibration sets, use the respective models to assign scores to each trajectory: 
    \begin{equation}
        \ell_{\text{success}}(\tau_t), \qquad \ell_{\text{fail}}(\tau_t).
    \end{equation}

    \item Use the scores to derive fixed thresholds $\eta_s$ and $\eta_f$ 
    using conformal prediction (CP). These thresholds specify cutoff values for anomalies for each decision function:
    \begin{align}
        D_{\text{success}}(\tau_t;\theta_s) = \mathbf{1}\big(\ell_{\text{success}}(\tau_t) > \eta_s\big), \\ 
        D_{\text{fail}}(\tau_t;\theta_f) = \mathbf{1}\big(\ell_{\text{fail}}(\tau_t) > \eta_f\big).
    \end{align}

    \item Combine the two detectors to obtain the consensus OOD decision using \Cref{eq:total_decision_fxn}.

\end{enumerate}

With its CP basis, this framework is policy-agnostic, distribution-free, and compatible with many scoring methods. 


\subsection{Classifying Failures for Gauge Detection}

We apply our framework to autonomous gauge inspection in both office and industrial sites. The site contains numerous gauges that vary in location, size, and shape. A trajectory is classified as:
\input{model_architecture}
\begin{itemize}
    \item \textbf{Success:} The robot centers and orients itself and its onboard pan-tilt-zoom (PTZ) correctly to obtain an accurate gauge reading.
    \item \textbf{Failure:} The robot produces an inaccurate gauge reading, such as when the gauge is partially or fully occluded or viewed at a poor angle. The gauge cannot be centered because of the limited PTZ configuration space. These can be dangerous if the robot misses a reading that could indicate abnormal equipment behavior or unsafe conditions.
    \item \textbf{OOD:} The robot is unable to produce a reading due to suboptimal but non-safety-critical conditions (e.g., shadows, glare, blurriness, poor lighting). These cases are not unsafe since no incorrect measurement is reported, but they should not be considered successful.
\end{itemize}
Examples of all of these cases can be seen in \Cref{fig:gauges}. 

\input{figures/gauge_classes}

The objective is to detect the earliest time $k \leq T$ at which a trajectory leads to failure, such that corrective actions can be taken early. The robot is tele-operated and assumes noise-free camera observations.

%% file: figures/decision_function.tex
\begin{figure}[t]
\vspace{2mm}
\centering
\resizebox{.98\columnwidth}{!}{%
\begin{tikzpicture}[
  >=Latex,
  node distance=8mm and 14mm,
  every node/.style={font=\small},
  block/.style={draw, rounded corners=2pt, align=center, text width=60mm, minimum height=8mm, very thick},
  leaf/.style={draw, rounded corners=2pt, align=center, text width=30mm, minimum height=6mm, very thick},
  arr/.style={->, very thick}
]

\node[block] (start) {Compute scores:\\
$\ell_{\text{success}}(\tau_t)$ from $M_{\text{success}}(\cdot;\phi_s)$, \quad 
$\ell_{\text{fail}}(\tau_t)$ from $M_{\text{fail}}(\cdot;\phi_f)$};

\node[block, below=10mm of start,
      minimum width=100mm] (check) 
{Evaluate thresholds simultaneously:\\
$\begin{gathered}
    D_{\text{success}}(\tau_t)=\mathbf{1}\!\big(\ell_{\text{success}}(\tau_t)>\eta_s\big) \\
    D_{\text{fail}}(\tau_t)=\mathbf{1}\!\big(\ell_{\text{fail}}(\tau_t)>\eta_f\big)
  \end{gathered}$};

\node[leaf, below left=12mm and 6mm of check] (succ) {\textbf{SUCCESS}\\($D_{\text{success}}{=}0,\;D_{\text{fail}}{=}1$)};
\node[leaf, below=16mm of check] (anom) {\textbf{ANOMALY}\\($D_{\text{success}}{=}1,\;D_{\text{fail}}{=}1$)};
\node[leaf, below right=12mm and 6mm of check] (fail) {\textbf{KNOWN FAILURE}\\($D_{\text{success}}{=}1,\;D_{\text{fail}}{=}0$)};

\draw[arr] (start) -- (check);

\draw[arr] (check.west) -| (succ.north)
  node[pos=0.25, above] {Only $D_{\text{fail}}{=}1$};

\draw[arr] (check.south) -- ++(0,-7mm)
  node[midway, right] {Both $=1$} -- (anom.north);

\draw[arr] (check.east) -| (fail.north)
  node[pos=0.25, above] {Only $D_{\text{success}}{=}1$};

\end{tikzpicture}%
}
\caption{Decision function for failure classification.}
\label{fig:decision_function}
\end{figure}
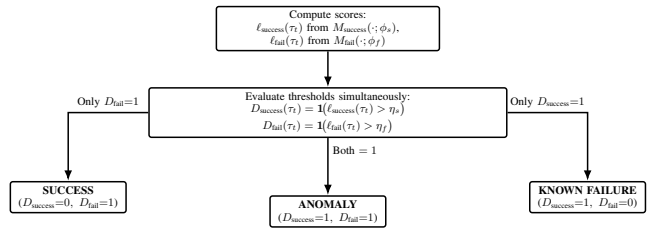

%% file: model_architecture.tex
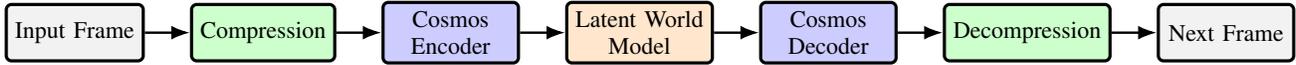
\begin{figure*}[t]
\vspace{2mm}
\centering
\begin{adjustbox}{max width=\textwidth}
  \begin{tikzpicture}[
    >=Latex,
    node distance=5mm and 6mm,
    every node/.style={font=\small},
    box/.style={draw, rounded corners=2pt, minimum height=7.5mm, minimum width=18mm, align=center, very thick, fill=gray!10},
    cosmos/.style={box, fill=blue!20},
    comp/.style={box, fill=green!20},
    world/.style={box, fill=orange!20},
    arr/.style={->, line width=0.9pt}
  ]
    \node[box]    (inputs)  {Input Frame};
    \node[comp,   right=of inputs] (comp)   {Compression};
    \node[cosmos, right=of comp]   (enc)    {Cosmos\\Encoder};
    \node[world,  right=of enc]    (world)  {Latent World\\Model};
    \node[cosmos, right=of world]  (dec)    {Cosmos\\Decoder};
    \node[comp,   right=of dec]    (decomp) {Decompression};
    \node[box,    right=of decomp] (outs)   {Next Frame};

    \draw[arr] (inputs) -- (comp);
    \draw[arr] (comp) -- (enc);
    \draw[arr] (enc) -- (world);
    \draw[arr] (world) -- (dec);
    \draw[arr] (dec) -- (decomp);
    \draw[arr] (decomp) -- (outs);
  \end{tikzpicture}
\end{adjustbox}
\caption{Model pipeline: Input frame from video feed is optionally compressed, tokenized by Cosmos, propagated through a latent world model, decoded by Cosmos, optionally decompressed, and reconstructed into the next frame.}
\label{fig:architecture}
\end{figure*}

%% file: figures/gauge_classes.tex
\begin{figure}[ht]
    \centering
    \begin{subfigure}{0.15\textwidth}
        \centering
        \includegraphics[width=\linewidth]{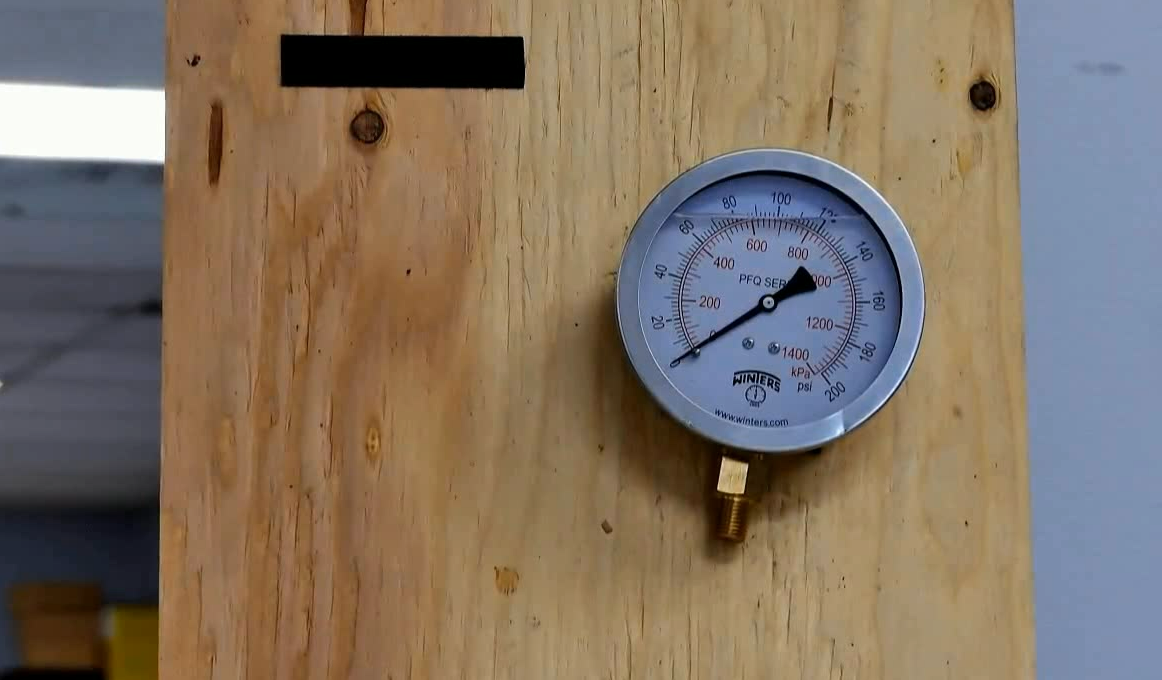}
        \caption{Success}
    \end{subfigure}
    \begin{subfigure}{0.15\textwidth}
        \centering
        \includegraphics[width=\linewidth]{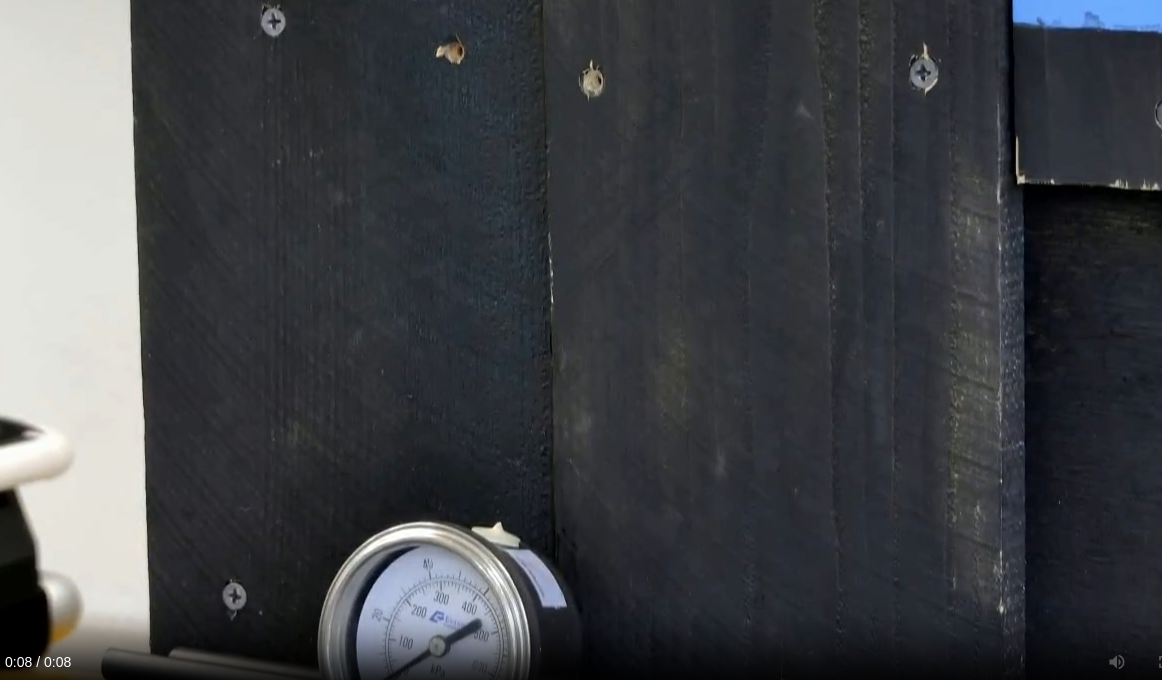}
        \caption{Failure}

    \end{subfigure}
    \begin{subfigure}{0.15\textwidth}
        \centering
        \includegraphics[width=\linewidth]{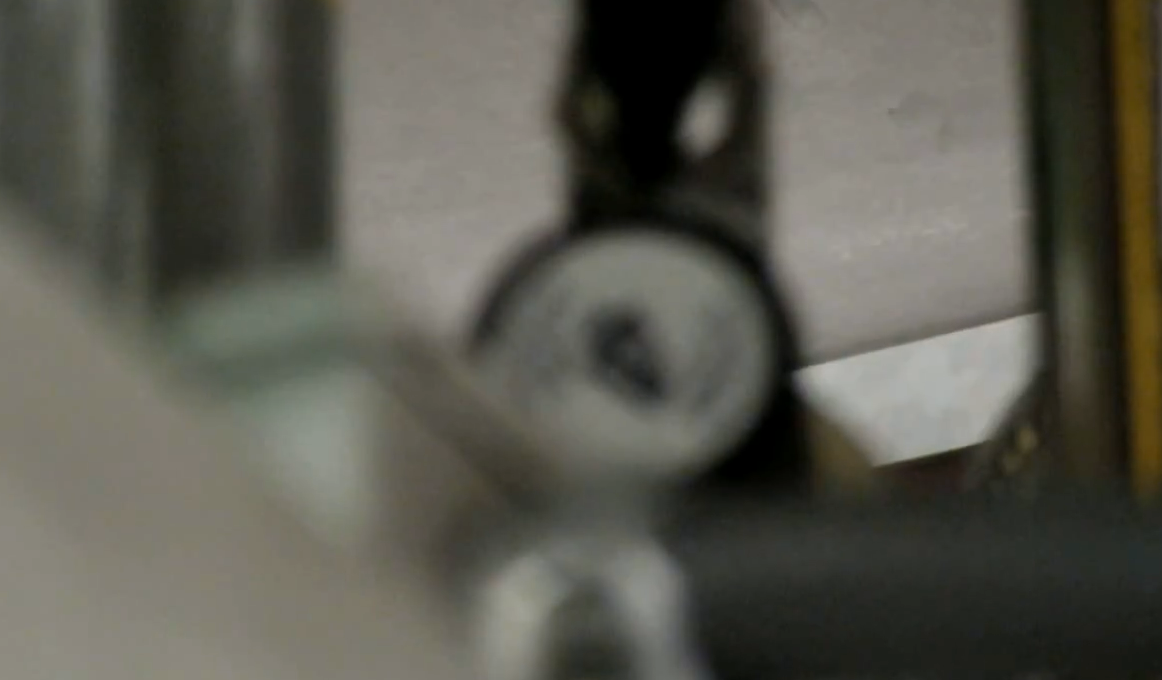}
        \caption{Anomaly (OOD)}
    \end{subfigure}

    \caption{Examples of Gauge Image Classifications. We assume a fixed robot pose, so only the PTZ camera is controlled.}
    \label{fig:gauges}
\end{figure}

%% file: methodology.tex
\subsection{Dataset Creation}
Since we needed to train and validate two models, calibrate two CP bands, and evaluate three classifications, we used the following dataset sizes: 

\begin{itemize}
    \item 14 success and 14 failure training videos 
    \item 6 success and 6 failure validation videos
    \item 45 success and 37 failure CP calibration videos
    \item 45 success, 37 failure, and 53 OOD test videos,
\end{itemize}
for a total of 290 videos, each on average 10 seconds long. This choice was inspired by prior work on encoding synchronized camera feeds with a world model~\cite{ward2025world}. The set of known failures includes missing the gauge in the field of view, partial obstruction of the gauge or its needle, and viewing it at a wide angle that prevents an accurate reading. The OOD dataset consists of conditions that prevent the robot from getting a gauge measurement, such as shadows, glare, and blurriness. Our dataset was limited to what was collected from July 9, 2025, to August 8, 2025, from our two environments, an office and an industrial site.

\subsection{Model Architecture and Training}

The model architecture is adapted from prior work~\cite{ward2025world}, which employed a world model trained on a success dataset from multiple synchronized camera feeds. In contrast, we extend the approach by requiring inference from two models along with their respective conformal prediction bands. Consistent with the prior work~\cite{ward2025world}, we use the NVIDIA Cosmos Tokenizer (Continuous Videos)~\cite{agarwal2025cosmos}, but keep the weights frozen. Its pretraining eliminated the need to train a large VAE ourselves. The model was used on video feeds to predict future frames. For a video feed, tokenized latents are concatenated along channels, yielding shape $[B,H/16,W/16]$. A latent world model then predicts $z_{t+1}$ from $z_t$. Before being fed into the Cosmos encoder, the images are compressed from $1200 \times 700$ to $512 \times 288$, to reduce memory usage and speed up training. The flow is illustrated in \Cref{fig:architecture}. 

Training is managed through the PyTorch Lightning Trainer API. \Cref{tab:train_params} summarizes the training parameters. We train the world model with a composite loss that balances frame reconstruction, temporal consistency, and anomaly–sensitive regularization:
\begin{equation}
    \mathcal{L}_{\text{total}}
    = \mathcal{L}_{\text{rec}}
    + \big(\mathcal{L}_{\text{rec}} - \mathcal{L}_{\text{cross}}\big)
    + 0.5\,\mathcal{L}_{\text{hyb}}.
\end{equation}

Here, $\mathcal{L}_{\text{rec}}$ is a weighted pixel loss (MSE+SSIM) between predicted and ground–truth next frames, $\mathcal{L}_{\text{cross}}$ enforces temporal consistency by encouraging predictions to be closer to the next frame than the current frame, and $\mathcal{L}_{\text{hyb}}$ combines latent prediction error, perceptual similarity (LPIPS), and a weak center–region prior. We use a 70/30 training/validation split. Because consecutive frames often contain minimal differences, we reduce redundancy by applying a skip parameter to lower computation time and overhead. We store per-frame latents $\{z_t, z_{t+1}\}$ to be used with some CP metrics. Training each model took about 34 hours with early stopping.



\subsection{Conformal Prediction Threshold Calibration}
\begin{figure}[h]
    \centering
    \includegraphics[width=0.6\linewidth]{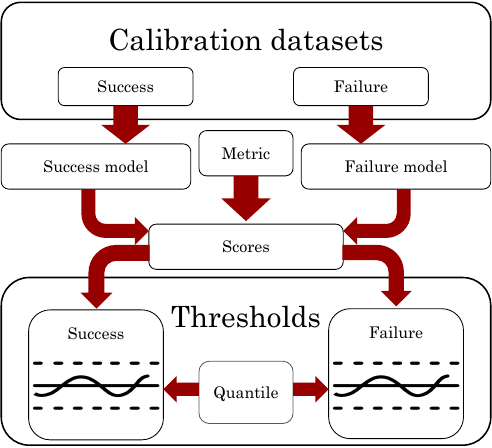}
    \caption{Threshold Calibration Phase}
    \label{fig:calibration}
\end{figure}
We adopt CP to calibrate thresholds on anomaly scores, consistent with recent work~\cite{xu_can_2025, ward2025world}. Our method enables classification of trajectories into success, known failure, or OOD. Each model is calibrated with its own calibrated band: one based on success data and one based on failure data. However, as acknowledged by previous work~\cite{ward2025world}, distribution shifts can arise from environmental conditions, operator behavior, or hardware degradation. Then, bands calibrated on earlier data may no longer reflect deployment conditions, leading to misestimated detection rates without adaptive recalibration.

\subsubsection{Calibration}
\begin{table*}[!t]
\vspace{2mm}
\centering
\caption{CP scoring methods, grouped by category, from CP literature \cite{angelopoulos2023conformal} and prior CP thresholding work \cite{ward2025world}. For all metrics in this table, values higher than the calibrated threshold indicate a potential out-of-distribution sample.}
\begin{tabular}{@{}l l p{11.5cm}@{}} 
\hline
\textbf{Method} & \textbf{Type} & \textbf{Description} \\
\hline
Reconstruction error & Residual-based & Pixel-level error between input and reconstruction. \\
Latent Prediction error     & Residual-based & Deviation between latent representations of predicted and actual future frames. \\
Latent standard deviation & Residual-based & Temporal standard deviation of latent embeddings. \\
\hline
Mahalanobis distance & Distance-based & Squared distance of latent embedding from the estimated Gaussian distribution. \\
Latent distance (L2) & Distance-based & Euclidean distance between latent embedding and calibration mean. \\
Latent cosine distance & Distance-based & Complement of cosine similarity (directional alignment of latent vectors). \\
\hline
Training loss        & Miscellaneous & Model’s per-sample training loss function. \\
\hline
\end{tabular}

\label{tab:scoringmethods}
\end{table*}

For each trained model, we create two conformal prediction bands from separate calibration sets of success and failure data. We use a maximum score per trajectory instead of timestep-level scoring, so that decisions are not affected by temporal correlations in a trajectory. A visual representation of this process is shown in \Cref{fig:calibration}. Consistent with previous CP anomaly detection work~\cite{ward2025world}, since we cannot assume CP validity guarantees without perfect calibration and test data exchangeability, we set thresholds at a chosen quantile level $(1-\alpha)$. Because thresholds are set below the maximum calibration trajectory score, some false negatives are expected, but this trade-off minimizes false positives, which is more critical for safe deployment. To assess metric suitability, we visually compare success and failure score distributions, where substantial overlap indicates poor separation. 

We test multiple CP residual and distance-based scoring metrics, as described in \Cref{tab:scoringmethods} from the CP literature \cite{angelopoulos2023conformal}. We cannot use action-dependent metrics \cite{xu_can_2025} due to tele-operation. We also compare against training loss since it indicates how well the model recognizes in-distribution data, but does not amplify the separation between that and OOD data. Since the models predict the next frame, we assign scores to frame pairs and summarize each trajectory by the maximum frame-pair score. For efficiency in real-time deployment, we adopt constant-value bands, similar to the prior work in failure detection with world models \cite{ward2025world}, rather than time-varying bands \cite{xu_can_2025}, and trim videos to equal length so no sample dominates calibration.

\subsubsection{Evaluation}
At test time, we score new videos drawn from success, failure, or OOD test sets and classify them using the decision function from \Cref{eq:total_decision_fxn}. 
If the distributions are separate, a success will only fall within the success band and outside the failure band, a failure will only fall within the failure band and outside the success band, and an OOD sample will fall outside of both bands. Calibrating each band and classifying one test trajectory data set takes $\sim$1.5 hours. We demonstrate our framework on both offline video streams taken from the robot as well as on hardware.


%% file: results.tex
We use classification accuracy and detection time as our success metrics, like Xu et al. \cite{xu_can_2025}, and evaluate our framework on various conformal prediction metrics and thresholding quantiles. We plot histograms to assess each metric by the separation between distributions.

\subsection{Detection Accuracy}
\input{tables/accuracy}
\input{figures/plots}

Among the results shown in \Cref{tab:accuracy}, Mahalanobis distance achieved the highest OOD detection accuracy (100\% across thresholds) but was prone to overfitting to the calibration set. In high-dimensional settings, covariance inversion can be unstable, which can lead to misclassifying valid but visually different successes as OOD. Regularization (shrinkage, PCA) reduced this instability but collapsed distinctions between nominal and OOD. Latent prediction error proved more reliable, achieving over 90\% accuracy across classes at the 90\% quantile. Unlike Mahalanobis, it does not rely on covariance inversion, instead measuring discrepancies between predicted and actual latent states. By contrast, L2 distance and latent standard deviation underperformed. L2 distance ignores covariance and temporal context, while standard deviation captures variability but not dynamic progression with time. Reconstruction error and latent cosine difference performed the worst. Since Cosmos was originally trained for reconstruction, it left little anomaly signal even for a failure video. Performance improved when inputs were more compressed (\Cref{tab:lowres-accuracy}). For the latent cosine difference, the high-dimensional latent space lacked a consistent dominant direction, so the average vector was poorly defined. Training loss classified the nominal data well, as expected. Since it is not designed to amplify anomalies, this metric resulted in weaker separation, notably at the 95\% threshold.

The histograms of score distributions in \Cref{fig:distributions} support the above results. Mahalanobis showed clean separation, while most showed overlap. This observation suggests the importance of quantile threshold choice. Because most metric distributions overlap, shifting the threshold inevitably misclassifies some portion of nominal or OOD data. Lower quantiles increased OOD accuracy but reduced nominal accuracy, while higher quantiles had the opposite effect. In safety-critical domains, false positives are preferable to false negatives. In our case, the 100\% quantile was overly conservative, while 90\% achieved the best balance, with latent prediction error classifying $\sim$91\% correctly across all three classes.

Finally, the success model consistently encompassed the detections made by the failure model, i.e., it classified all the same OOD cases. This pattern likely occurred since the failure model was exposed to diverse failure types, while the success model was trained on a single, more consistent case, yielding better separation.

\subsection{Detection Time}
Another success metric that we considered was the difference between the true failure detection time, as observed by a human, as opposed to the time at which the framework made a classification. The results can be seen in \Cref{tab:detection_times}.

\input{tables/times}

Nearly all metrics were able to predict the class before they became apparent to a human observer. This pattern indicates that the metrics are sensitive to underlying visual or latent patterns that precede observable failure modes. The most successful metric for accuracy, latent prediction error, made classifications on average between 1 and 3 seconds earlier than a human observer and almost consistently had the lowest standard error. An example run over a 10-second period can be seen in \Cref{fig:detection_bands}. Mahalanobis was the earliest, but since it is prone to overfitting, it reduces interpretability and increases the likelihood of false positives. The training loss was the only metric that detected failures later than the human observer, since the loss function amplifies nominal patterns and lacks any contrastive signal for anomalies.

\input{figures/detection_bands}

\subsection{Hardware Results}

We validate our framework on a hardware platform, seen in \Cref{fig:detection_bands}, demonstrating its ability to operate in real time. We deployed a Boston Dynamics Spot, equipped with LiDAR, cameras for navigation \cite{bouman2020autonomous}, and a Pan-Tilt-Zoom (PTZ) camera for detecting gauges, which moves independently of the robot's pose. Due to shared-use restrictions, we were unable to host the models directly on board the robot. Instead, inference was hosted externally. However, the full pipeline only requires at most $\sim$1.5 GB of storage, dominated by the pretrained Cosmos weights ($\sim$1.3 GB), with our class-specific models ($\sim$60 MB) and auxiliary CP artifacts (thresholds, precomputed values, $<$ 5 MB), so we suspect it can be deployed onboard. We experimented with several gauges placed in an office.

At runtime, the PTZ camera captures two frames and processes them through the framework (\Cref{fig:runtime_classification}). In the case of a success, the robot remains idle until commanded to proceed to the next gauge. For a failure, the system records the affected gauge, which will be revisited after all gauges have been surveyed. For OOD inputs due to a distant, blurry reading, the robot initiates a zoom operation and reprocesses the new frames through the pipeline. A single classification run requires approximately three minutes, primarily due to network latency from passing data to the externally hosted models. The actual scoring is comparably instantaneous when using a precomputed threshold. While this latency is acceptable in the current application, higher-risk tasks would benefit from hosting the model locally to enable immediate detection and intervention. Nevertheless, we designed our framework to employ lightweight models, observation-based metrics that do not require auxiliary networks, and constant-valued thresholds, successfully demonstrating its abilities on a real hardware platform.

%% file: tables/accuracy.tex
\begin{table}[t]
\centering
\scriptsize
\renewcommand{\arraystretch}{1.05}
\caption{Detection accuracy (\%) of all three classes: success, known failure, and OOD, for three different quantiles. Three models were trained for both success and failure. Detection rates were consistent, except the reconstruction error for detecting failures improved by one trajectory.}
\begin{subtable}{\columnwidth}
\centering
\setlength\tabcolsep{2pt}
\begin{tabular}{@{}lccccc@{}}
\hline
Metric & OOD (Succ) & OOD (Fail) & OOD (Total) & Fail (Total) & Succ (Total) \\
\hline
Reconstruct. error     & 52.83 & 49.06 & 49.06 & 48.65 & 40.00 \\
Latent pred. error  & 94.34 & 90.57 & \textbf{90.57} & \textbf{91.89} & \textbf{91.11} \\
Latent std dev.           & 64.15 & 52.83 & 52.83 & 59.46 & 33.33 \\
Mahalanobis              & 100.00 & 100.00 & 100.00 & 90.00 & 90.00 \\
Latent dist. (L2)                       & 75.47 & 54.72 & 54.72 & 81.08 & 40.00 \\
Latent cosine dist.                   & 45.28 & 43.40 & 39.62 & 67.57 & 42.22 \\
Training Loss            & 90.57 & 84.91 & 84.91 & 89.19 & 80.00 \\
\hline
\end{tabular}
\caption{90\% threshold}
\end{subtable}

\vspace{0.35em}

\begin{subtable}{\columnwidth}
\centering
\setlength\tabcolsep{2pt}
\begin{tabular}{@{}lccccc@{}}
\hline
Metric & OOD (Succ) & OOD (Fail) & OOD (Total) & Fail (Total) & Succ (Total) \\
\hline
Reconstruct. error     & 35.85 & 45.28 & 35.85 & 18.92 & 37.78 \\
Latent pred. error  & 86.79 & 77.36 & \textbf{77.36} & \textbf{89.19} & \textbf{75.56} \\
Latent std dev.           & 56.60 & 30.19 & 30.19 & 43.24 & 6.67 \\
Mahalanobis              & 100.00 & 100.00 & 100.00 & 95.00 & 95.00 \\
Latent dist. (L2)                       & 64.15 & 37.74 & 37.74 & 72.97 & 17.78 \\
Latent cosine dist.                   & 39.62 & 35.85 & 35.85 & 51.35 & 28.89 \\
Training Loss            & 75.47 & 54.72 & 54.72 & 75.68 & 51.11 \\
\hline
\end{tabular}
\caption{95\% threshold}
\end{subtable}

\vspace{0.35em}

\begin{subtable}{\columnwidth}
\centering
\setlength\tabcolsep{2pt}
\begin{tabular}{@{}lccccc@{}}
\hline
Metric & OOD (Succ) & OOD (Fail) & OOD (Total) & Fail (Total) & Succ (Total) \\
\hline
Reconstruct. error     & 18.87 & 16.98 & 16.98 & 16.98 & 0.00 \\
Latent pred. error  & 0.00  & 7.55  & 0.00  & 0.00  & 8.89 \\
Latent std dev.           & 0.00  & 2.70  & 0.00  & 2.70  & 0.00 \\
Mahalanobis              & 100.00 & 100.00 & 100.00 & 100.00 & 100.00 \\
Latent dist. (L2)                       & 0.00  & 0.00  & 0.00  & 0.00  & 4.44 \\
Latent cosine dist.                   & 0.00  & 0.00  & 0.00  & 0.00  & 6.67 \\
Training Loss            & 0.00 & 9.43 & 0.00 & 0 & 7.55 \\
\hline
\end{tabular}
\caption{100\% threshold}
\end{subtable}
\label{tab:accuracy}
\end{table}

%% file: figures/plots.tex
\begin{figure*}[t] 
    \centering
    \begin{subfigure}{0.25\textwidth}
        \includegraphics[width=\linewidth]{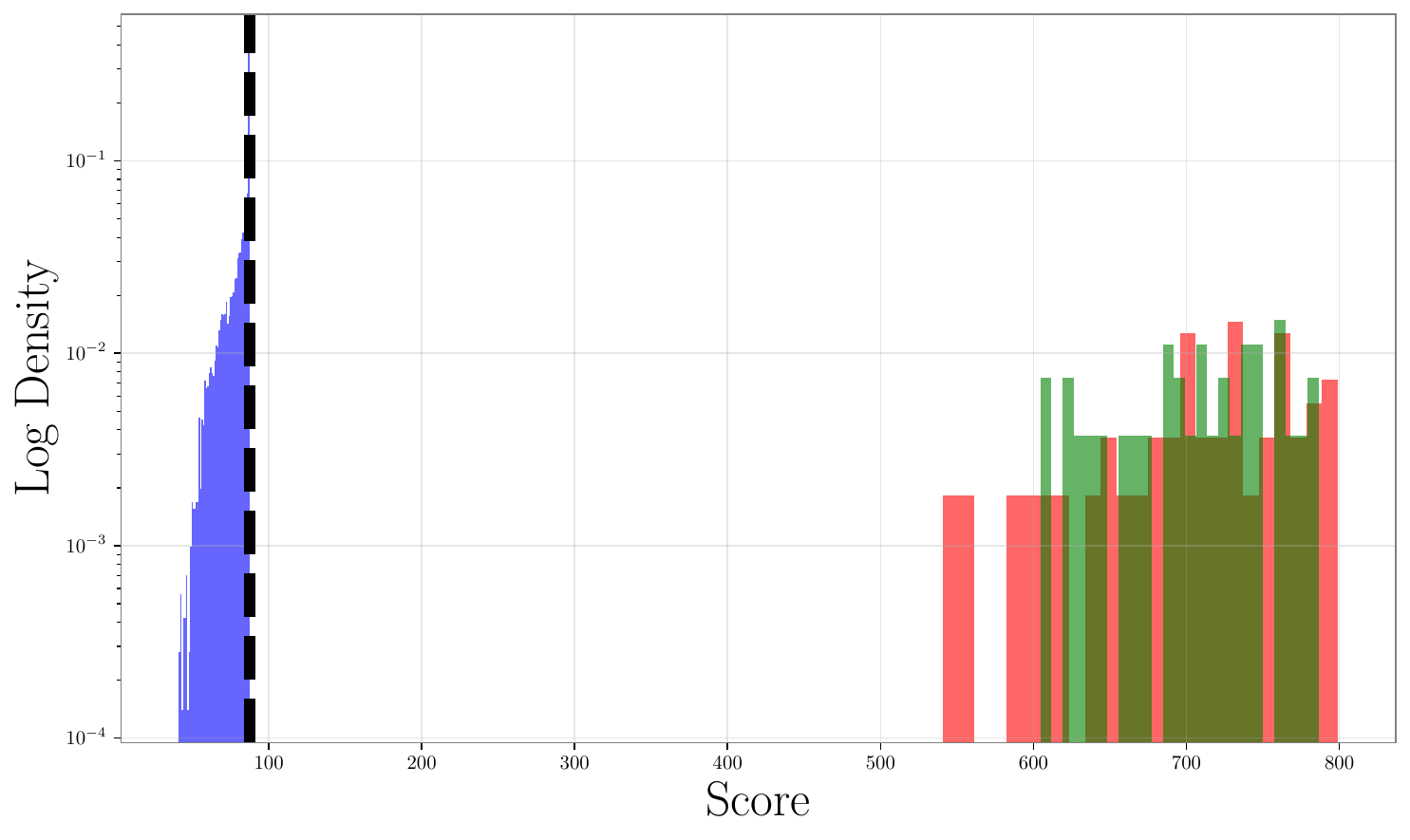}
        \caption{Mahalanobis Success}
    \end{subfigure}\hfill
    \begin{subfigure}{0.25\textwidth}
        \includegraphics[width=\linewidth]{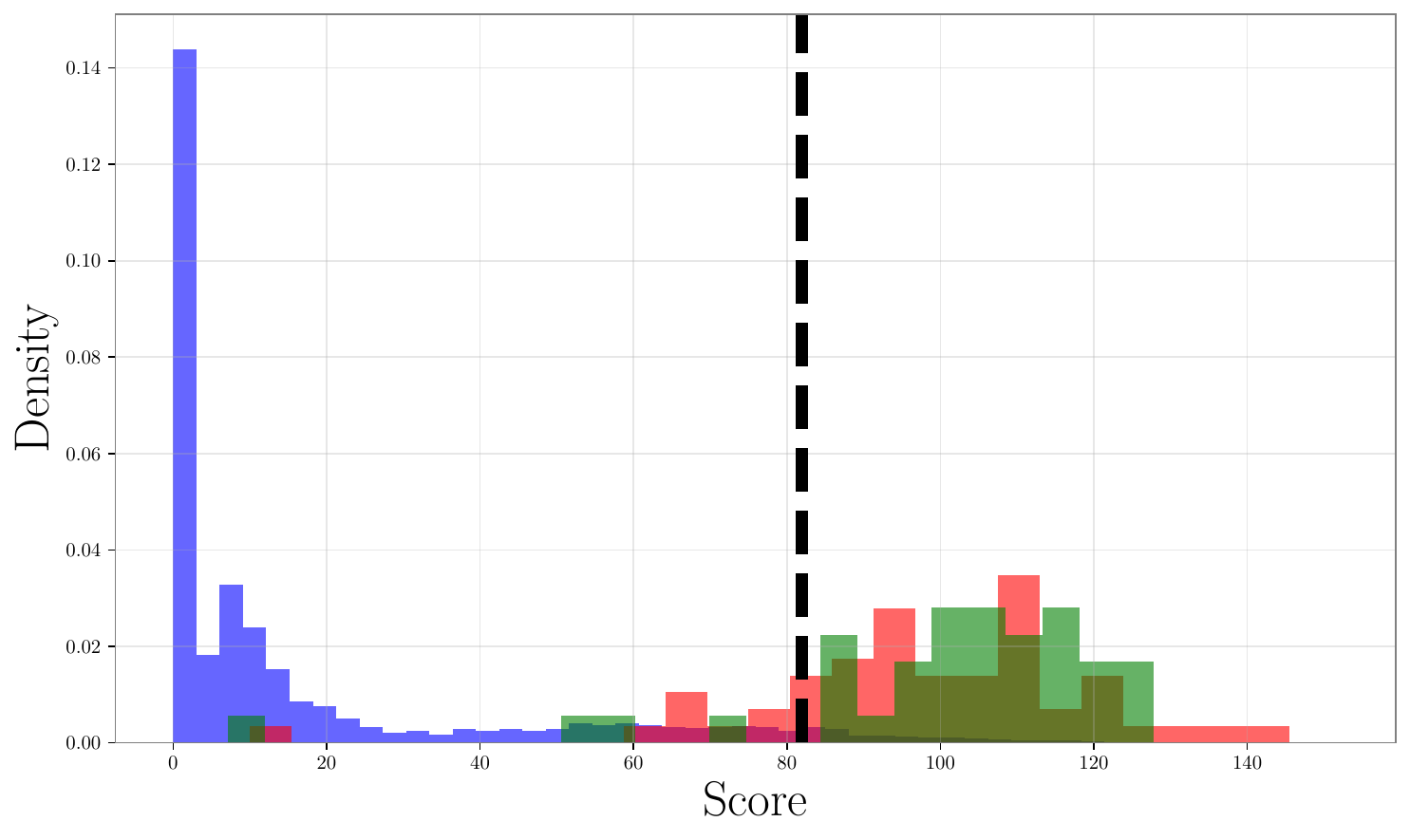}
        \caption{Latent Pred. Error Success}
    \end{subfigure}\hfill
    \begin{subfigure}{0.25\textwidth}
        \includegraphics[width=\linewidth]{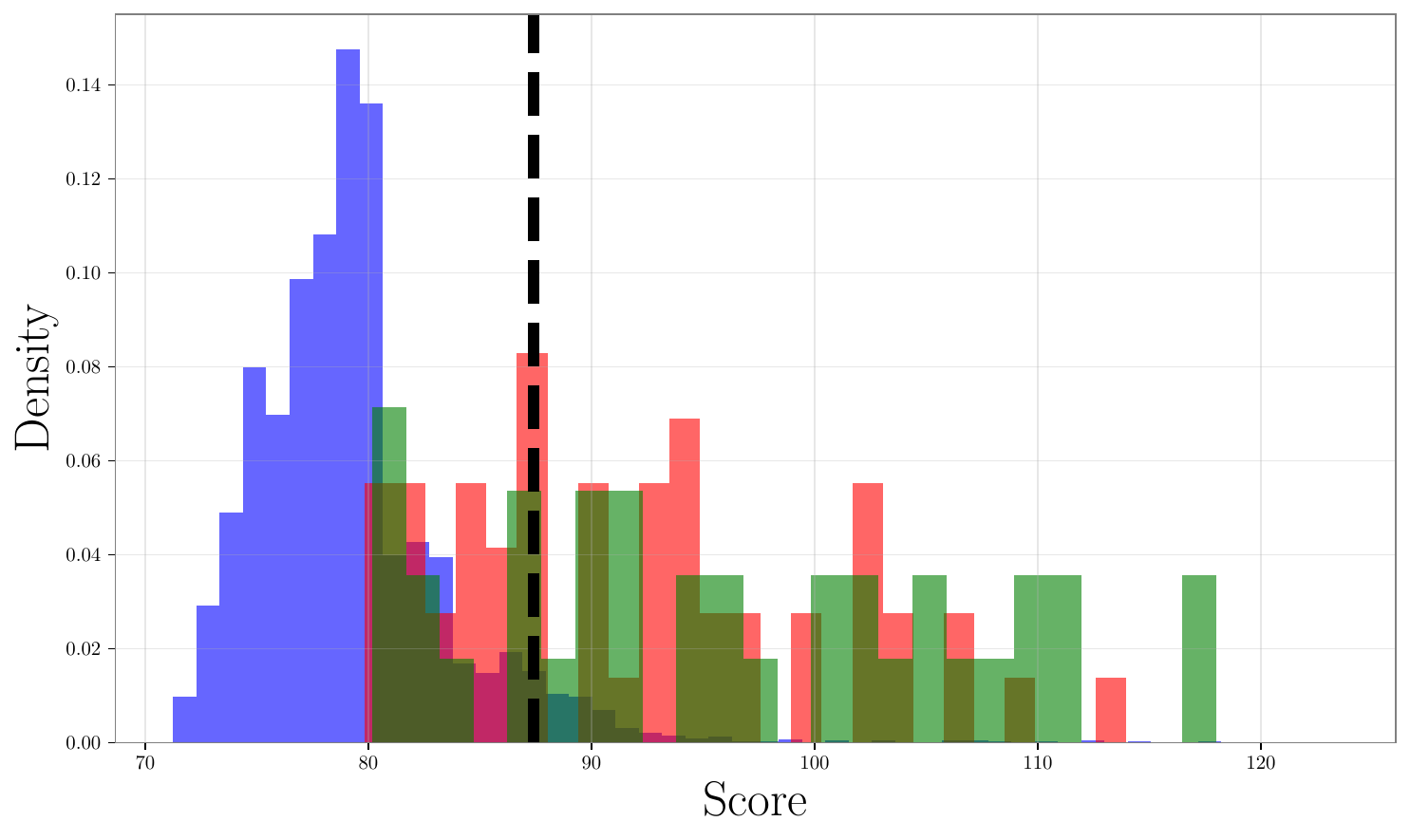}
        \caption{Latent L2 Distance Success}
    \end{subfigure}\hfill
    \begin{subfigure}{0.25\textwidth}
        \includegraphics[width=\linewidth]{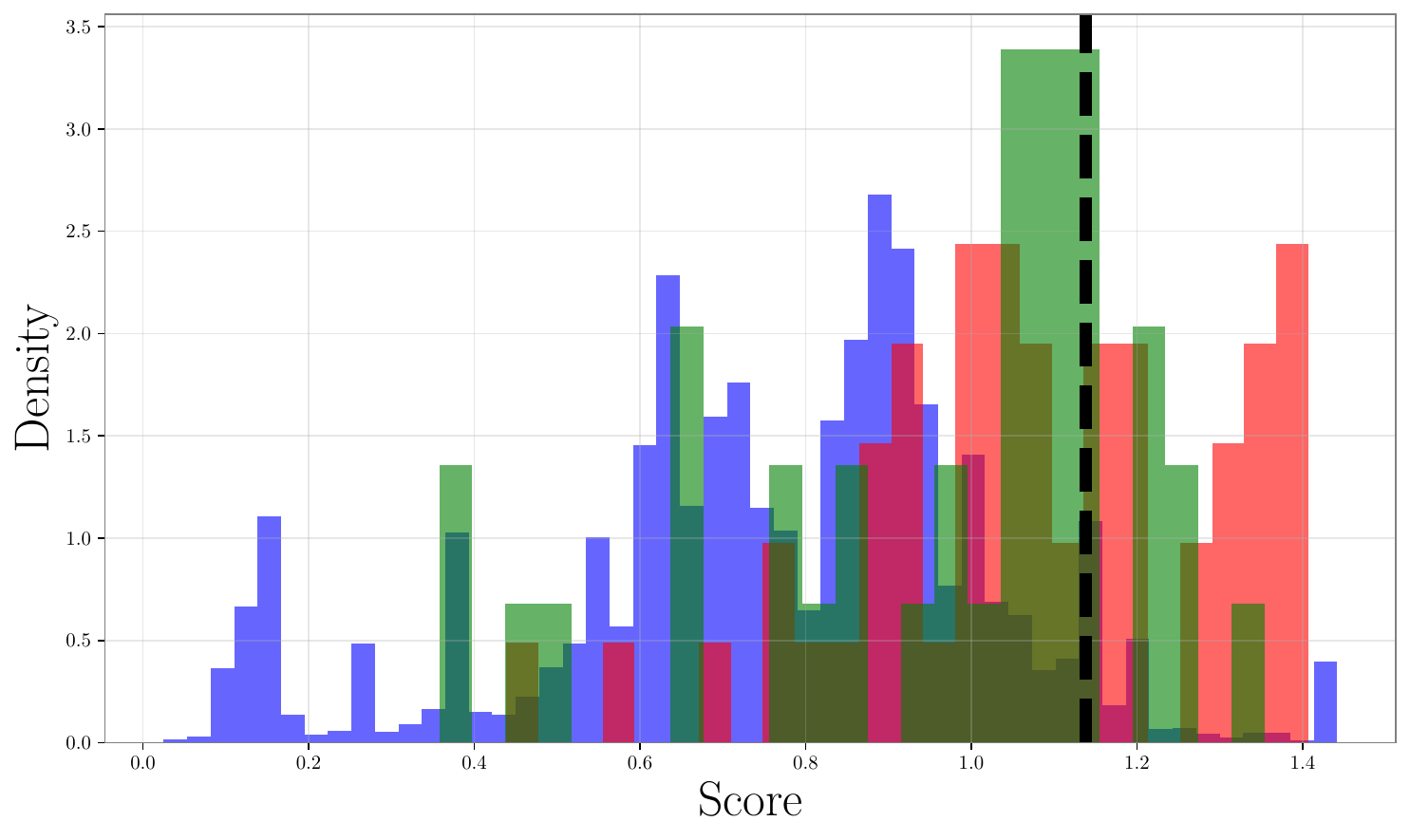}
        \caption{Reconstruction Success}
    \end{subfigure}

    \vspace{0.5em} 

    \begin{subfigure}{0.25\textwidth}
        \includegraphics[width=\linewidth]{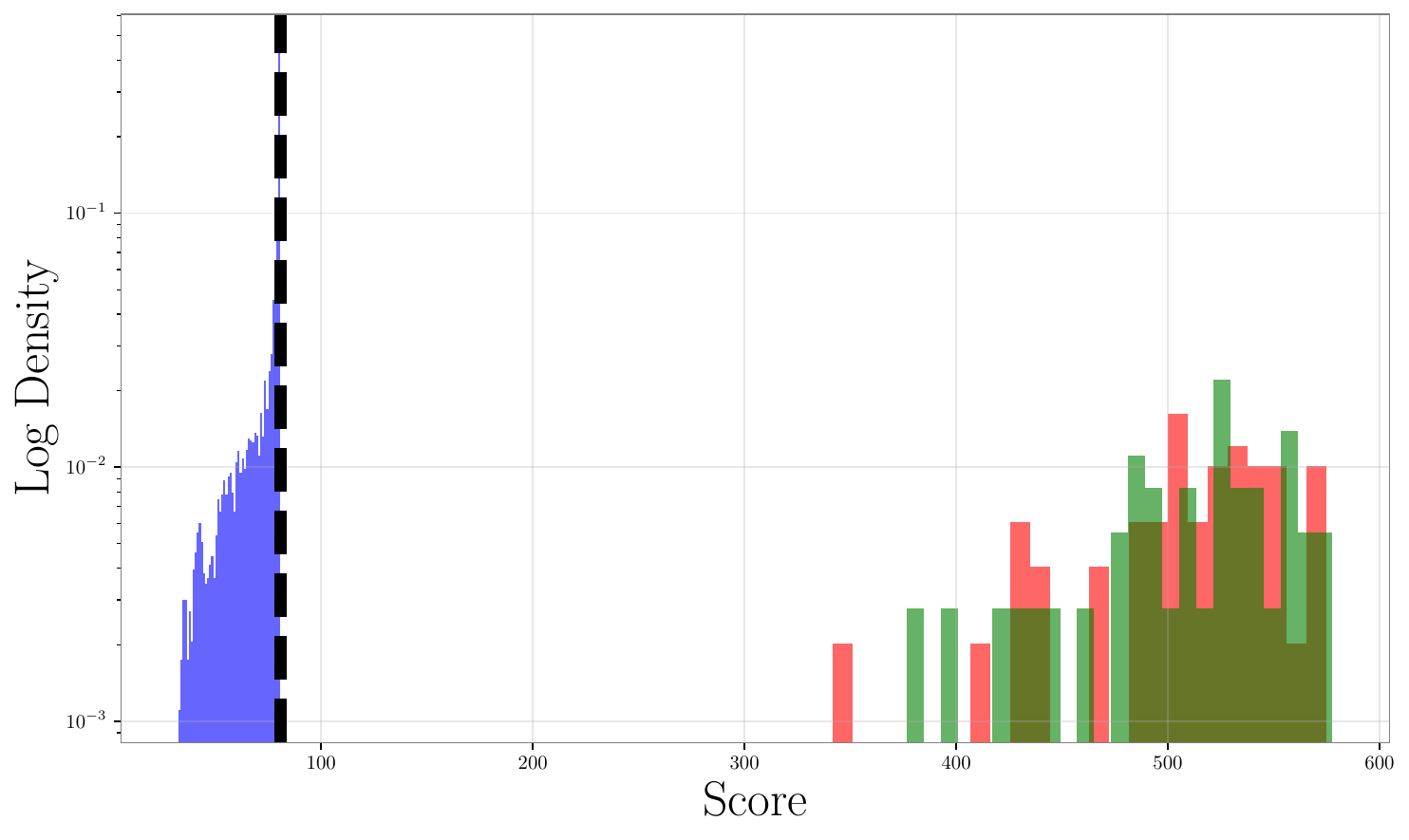}
        \caption{Mahalanobis Failure}
    \end{subfigure}\hfill
    \begin{subfigure}{0.25\textwidth}
        \includegraphics[width=\linewidth]{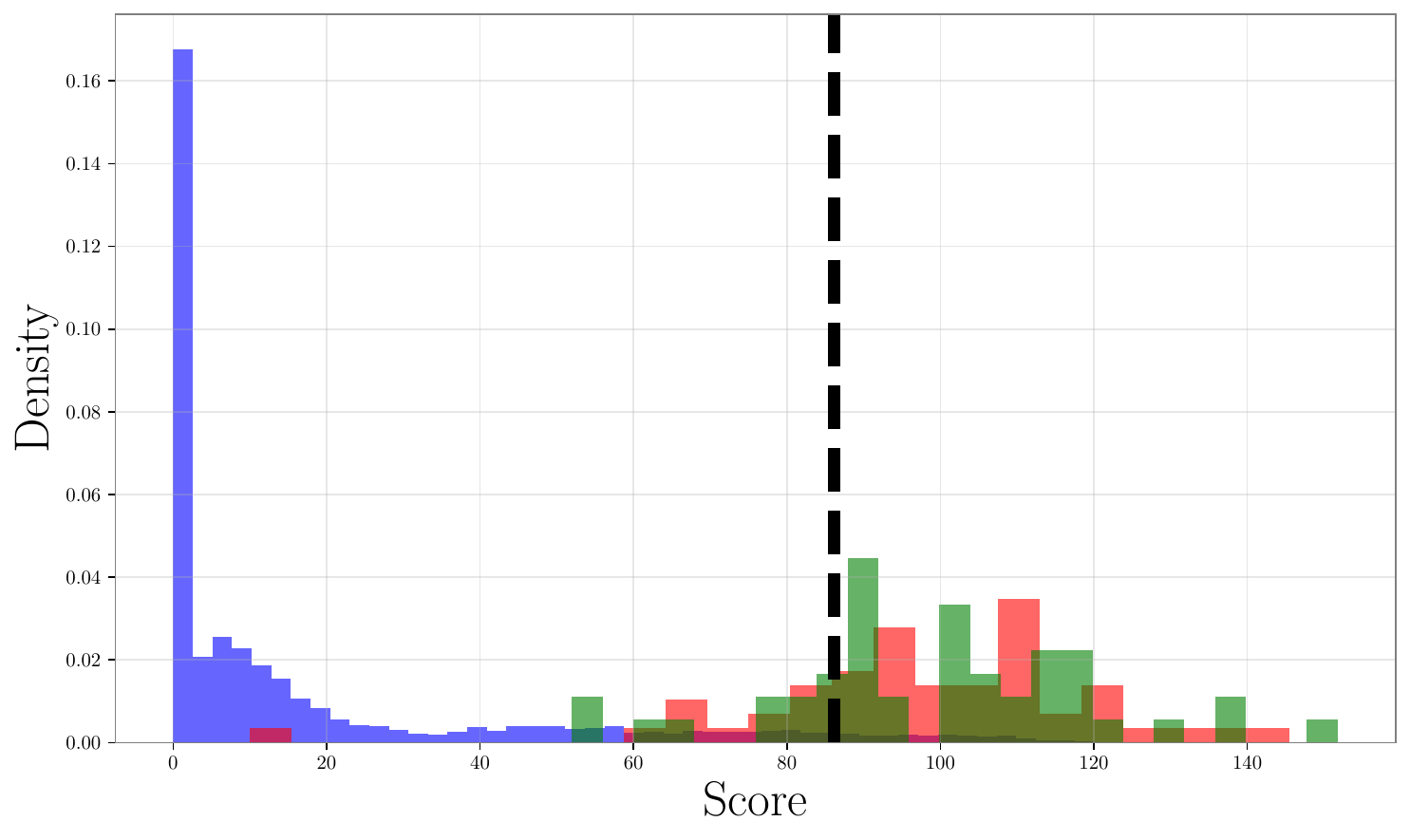}
        \caption{Latent Pred. Error Failure}
    \end{subfigure}\hfill
    \begin{subfigure}{0.25\textwidth}
        \includegraphics[width=\linewidth]{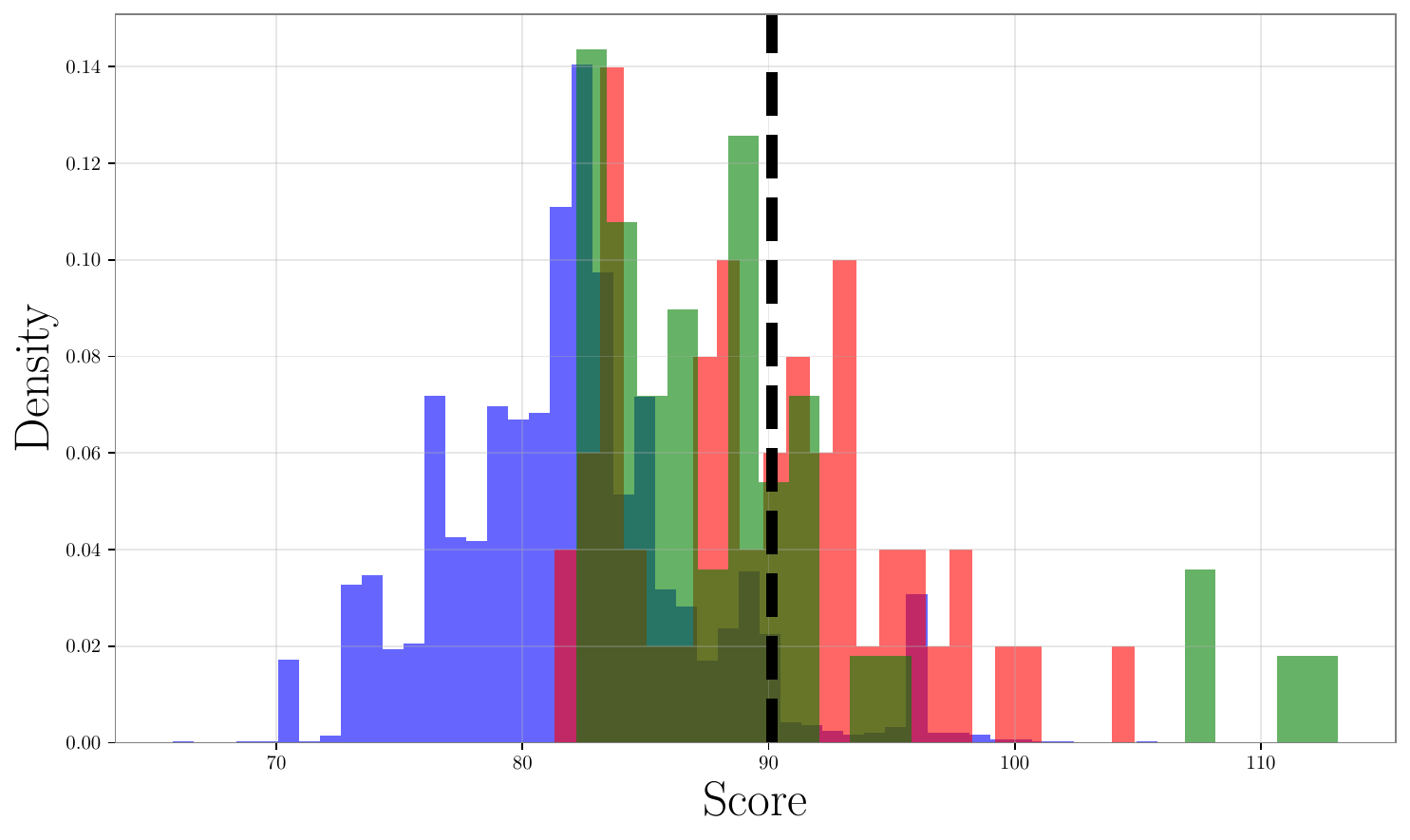}
        \caption{Latent L2 Distance Failure}
    \end{subfigure}\hfill
    \begin{subfigure}{0.25\textwidth}
        \includegraphics[width=\linewidth]{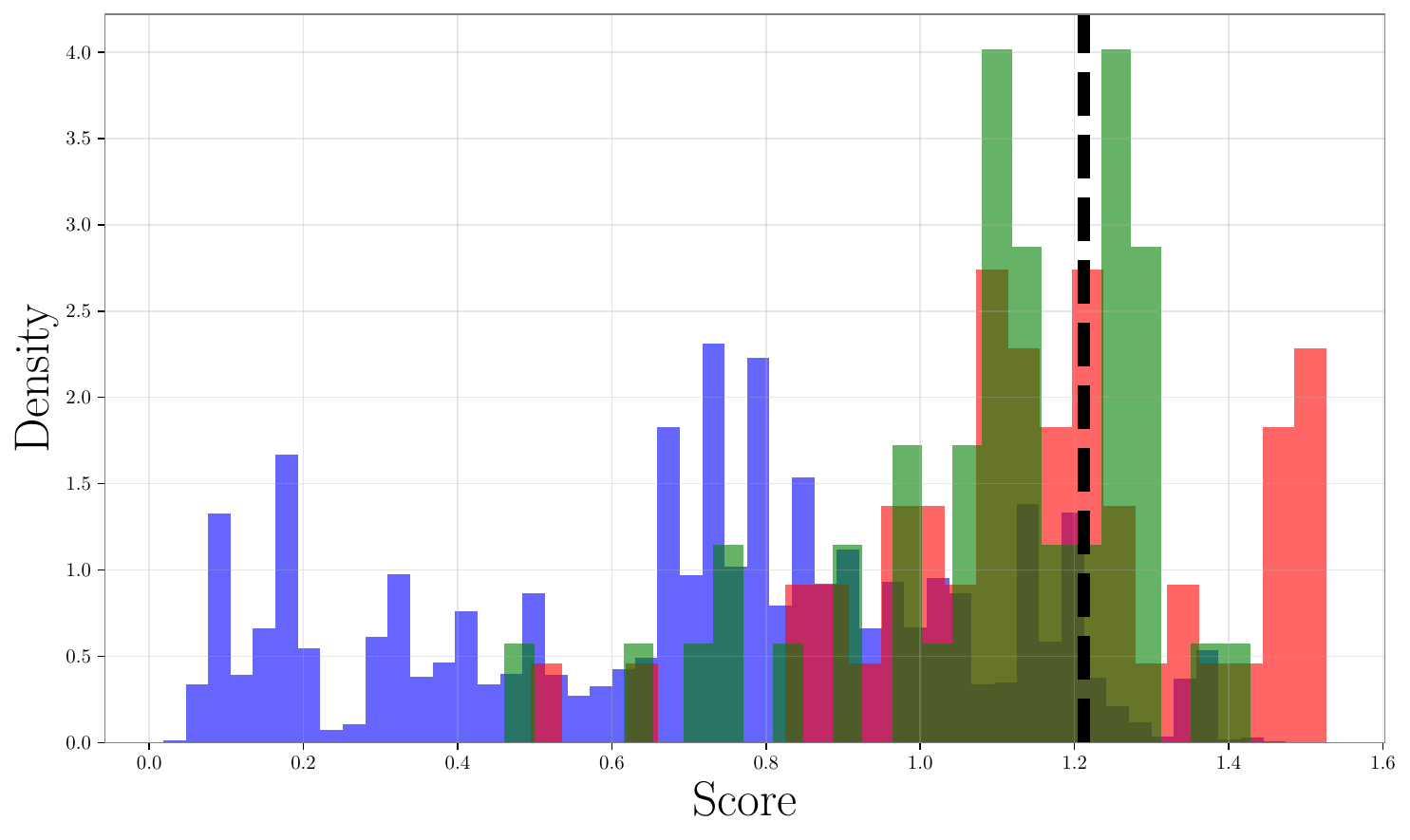}
        \caption{Reconstruction Failure}
    \end{subfigure}

    \caption{Nominal vs. OOD distributions scored by success and failure models on select metrics, with the 95\% quantile threshold shown with the black dashed line. Legend: (\textcolor{blue}{Nominal}, \textcolor{red}{OOD}, \textcolor{green}{Succ/Fail - Opposite Nominal})}
    \label{fig:distributions}
\end{figure*}

%% file: tables/times.tex
\begin{table}[h!]
\centering
\caption{Difference in detection time ($\hat{k}$), for each class given the model used. Results are evaluated at the 95\% quantile and presented as mean detection time in seconds $\pm$ standard error.}
\resizebox{\columnwidth}{!}{%
\begin{tabular}{@{}lcccc@{}}
\hline
Metric & OOD (Succ) & OOD (Fail) & Success (Fail) & Failure (Succ) \\
\hline
Reconstruct. error            & $-2.059 \pm 0.583$ & $-2.133 \pm 0.611$ & $-2.570 \pm 0.692$ & $-1.648 \pm 0.887$ \\
Latent pred. error           & $-1.844 \pm 0.406$ & $-2.517 \pm 0.490$ & $-3.526 \pm 0.452$ & $-3.440 \pm 0.533$ \\
Latent std dev.            & $-1.313 \pm 0.554$ & $-2.247 \pm 0.838$ & $-3.949 \pm 0.891$ & $-2.728 \pm 0.874$ \\
Mahalanobis    & $-5.974 \pm 0.438$ & $-5.974 \pm 0.438$ & $-9.005 \pm 0.191$ & $-8.637 \pm 0.618$ \\
Latent dist. (L2)       & $-1.686 \pm 0.565$ & $-2.669 \pm 0.796$ & $-3.737 \pm 0.679$ & $-2.217 \pm 0.807$ \\
Latent cosine dist.        & $-3.536 \pm 0.697$ & $-4.052 \pm 0.848$ & $-1.802 \pm 0.855$ & $-1.968 \pm 0.790$ \\
Training Loss  & $1.949 \pm 1.163$  & $2.090 \pm 1.054$  & $0.663 \pm 1.552$  & $0.934 \pm 1.115$  \\
\hline
\end{tabular}
}

\label{tab:detection_times}
\end{table}

%% file: figures/detection_bands.tex
\begin{figure*}[t]
    \centering
    \begin{subfigure}[t]{0.32\textwidth}
        \centering
        \includegraphics[width=\linewidth]{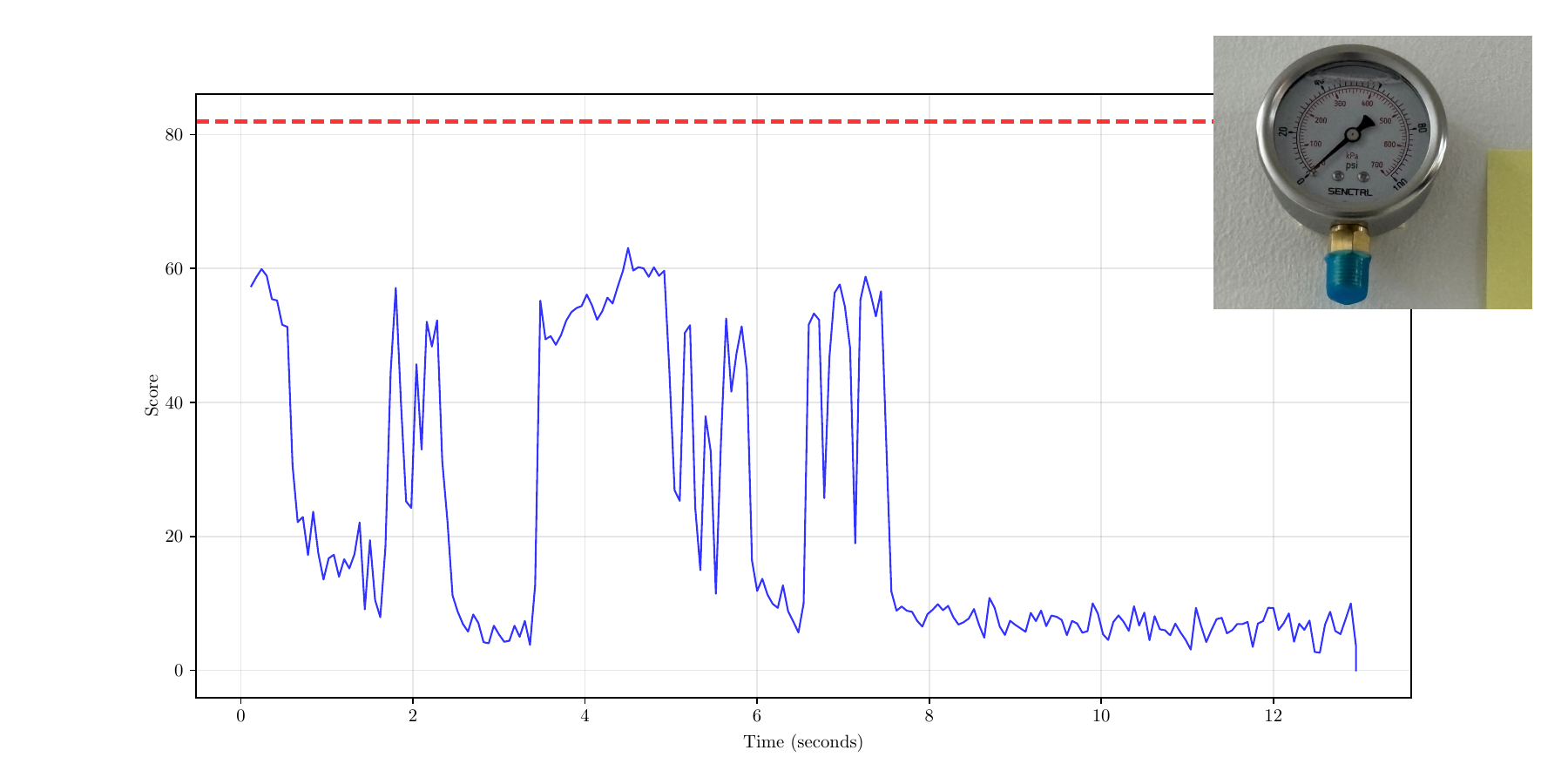}
        \caption{Succ. Detection with Succ. Model}
    \end{subfigure}
    \begin{subfigure}[t]{0.27\textwidth}
        \centering
        \includegraphics[width=\linewidth]{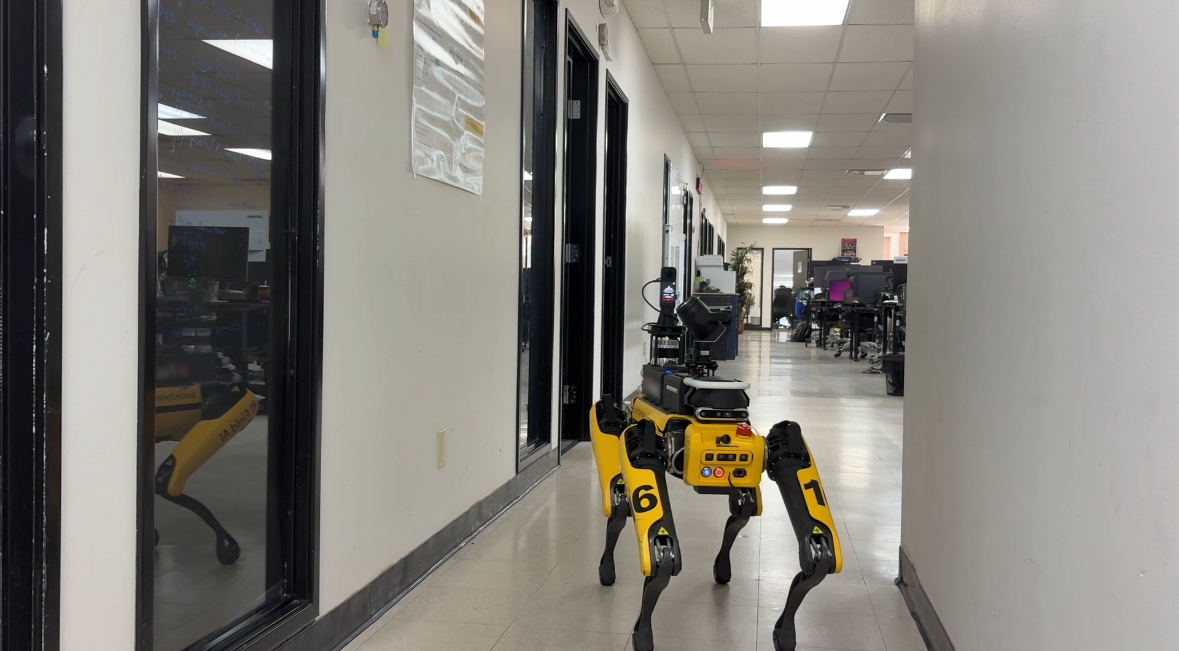}
        \label{fig:hardware_demo}
    \end{subfigure}
    \begin{subfigure}[t]{0.32\textwidth}
        \centering
        \includegraphics[width=\linewidth]{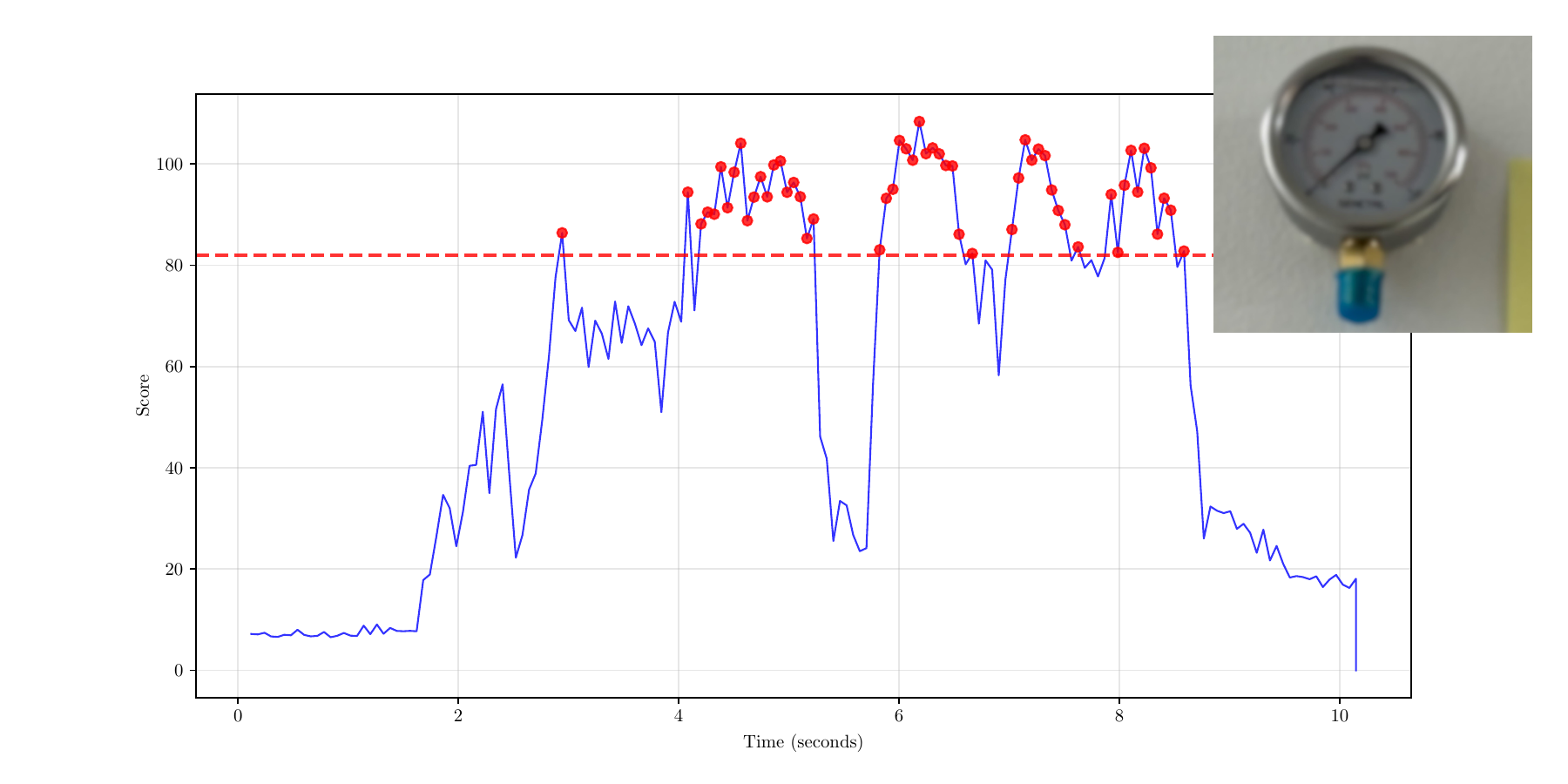}
        \caption{OOD Detection with Succ. Model}
    \end{subfigure}

    \caption{Hardware demonstration of framework for example success and OOD cases. Using latent prediction error, the scores were tracked with respect to the threshold over 10 seconds. The OOD case exceeds the threshold as expected.}
    \label{fig:detection_bands}
\end{figure*}

%% file: conclusion.tex






We presented a framework for anomaly detection and failure classification in the context of industrial gauge inspection using a world model backbone and conformal prediction thresholding. Our experiments demonstrated that we can separate between success, failure, and OOD classes at over 90\% accuracy, earlier than a human observer, highlighting the potential of the method for robust runtime monitoring for real-time inspection tasks. Beyond gauge inspection, such a framework can have a broader impact in enabling safer, more reliable deployment of autonomous systems in safety-critical industrial domains and identifying gaps in training data to guide targeted collection and retraining.

Our work has several opportunities for future improvement. We can address the distribution shifts between the calibration and test data with martingales~\cite{luo2024online}. Our current CP bands are static, and developing adaptive, time-varying bands could improve responsiveness to temporal drift. We only explored manual dimensionality reduction, so leveraging compression modules like singular value decomposition may enhance efficiency and accuracy. Incorporating semantic failure detection and more history could enable recognition of higher-level failures that reflect human-like understanding. Finally, training a single model with a classification head, rather than storing multiple models, can further decrease the computational overhead required to run the detector online.

%% file: tables/lowres_accuracy.tex
\begin{table}[h!]
\vspace{2mm}
\centering
\scriptsize
\renewcommand{\arraystretch}{1.05}
\caption{Detection accuracy (\%) across thresholds at 50\% higher image compression ($1200 \times 700$ to $256 \times 144$).}
\begin{subtable}{\columnwidth}
\centering
\setlength\tabcolsep{2pt}
\begin{tabular}{@{}lccccc@{}}
\hline
Metric & OOD (Succ) & OOD (Fail) & OOD (Total) & Fail (Total) & Succ (Total) \\
\hline
Reconstruct. error     & 60.38 & 49.06 & 45.28 & 51.35 & 42.22 \\
Latent pred. error  & 96.23 & 94.34 & 94.34 & 91.89 & 91.11 \\
Latent std dev.           & 64.15 & 52.83 & 52.83 & 59.46 & 17.78 \\
Mahalanobis              & 100.00 & 100.00 & 100.00 & 100.00 & 90.00 \\
Latent dist. (L2)                       & 75.47 & 67.92 & 66.04 & 81.08 & 57.78 \\
Latent cosine dist.                   & 52.83 & 24.53 & 24.53 & 64.86 & 20.00 \\

\hline
\end{tabular}
\caption{90\% threshold}
\end{subtable}
\hfill

\begin{subtable}{\columnwidth}
\centering
\setlength\tabcolsep{2pt}
\begin{tabular}{@{}lccccc@{}}
\hline
Metric & OOD (Succ) & OOD (Fail) & OOD (Total) & Fail (Total) & Succ (Total) \\
\hline
Reconstruct. error     & 39.62 & 41.51 & 33.96 & 32.43 & 35.56 \\
Latent pred. error  & 88.68 & 77.36 & 77.36 & 89.19 & 80.00 \\
Latent std dev.           & 49.06 & 30.19 & 30.19 & 43.24 & 17.78 \\
Mahalanobis              & 100.00 & 100.00 & 100.00 & 100.00 & 95.00 \\
Latent dist. (L2)                       & 60.38 & 32.08 & 32.08 & 72.97 & 13.33 \\
Latent cosine dist.                   & 35.85 & 20.75 & 20.75 & 37.25 & 13.33 \\

\hline
\end{tabular}
\caption{95\% threshold}
\end{subtable}
\hfill

\begin{subtable}{\columnwidth}
\centering
\setlength\tabcolsep{2pt}
\begin{tabular}{@{}lccccc@{}}
\hline
Metric & OOD (Succ) & OOD (Fail) & OOD (Total) & Fail (Total) & Succ (Total) \\
\hline
Reconstruct. error     & 7.55 & 22.64 & 7.55 & 2.70 & 6.67 \\
Latent pred. error  & 0.00 & 7.55 & 0.00 & 0.00 & 8.89 \\
Latent std dev.           & 0.00 & 0.00 & 0.00 & 2.70 & 0.00 \\
Mahalanobis              & 100.00 & 100.00 & 100.00 & 100.00 & 100.00 \\
Latent dist. (L2)                       & 0.00 & 0.00 & 0.00 & 0.00 & 6.67 \\
Latent cosine dist.                   & 0.00 & 0.00 & 0.00 & 2.70 & 0.00 \\

\hline
\end{tabular}
\caption{100\% threshold}
\end{subtable}
\label{tab:lowres-accuracy}
\end{table}